# Hybrid Emotion Recognition: Enhancing Customer Interactions Through Acoustic and Textual Analysis

Sahan Hewage Wewelwala
*School of Computing*
*Informatics Institute of Technology*
Colombo 06, Sri Lanka
sahanwewelwala@gmail.com

T.G.D.K. Sumanathilaka
*Department of Computer Science*
*Swansea University*
Swansea, Wales, United Kingdom
deshankoshala@gmail.com

*Abstract*—This research presents a hybrid emotion recognition system integrating advanced Deep Learning, Natural Language Processing (NLP), and Large Language Models (LLMs) to analyze audio and textual data for enhancing customer interactions in contact centers. By combining acoustic features with textual sentiment analysis, the system achieves nuanced emotion detection, addressing the limitations of traditional approaches in understanding complex emotional states. Leveraging LSTM and CNN models for audio analysis and DistilBERT for textual evaluation, the methodology accommodates linguistic and cultural variations while ensuring real-time processing. Rigorous testing on diverse datasets demonstrates the system's robustness and accuracy, highlighting its potential to transform customer service by enabling personalized, empathetic interactions and improving operational efficiency. This research establishes a foundation for more intelligent and human-centric digital communication, redefining customer service standards.

## I. INTRODUCTION

The capacity to identify and comprehend emotions effectively is an essential element of human-computer interaction, especially in spoken and written communication. Misinterpretation of emotional signals can result in a deficiency of comprehension, adversely affecting user experience and decision-making processes [1]. Speech emotion recognition (SER) systems have difficulties in accurately capturing the subtleties of emotions conveyed through variations in tone, pitch, and context. Overcoming these issues requires the creation of sophisticated computer techniques capable of successfully analyzing both auditory and language characteristics.

Recent advancements in deep learning (DL) and natural language processing have yielded effective approaches for emotion identification. Research has utilized hybrid systems that integrate audio data with language analysis to enhance precision and contextual comprehension [2]. These methodologies employ pre-trained models like BERT, ResNet, and other transformer-based architectures, which have shown effective in analyzing emotions across various datasets [3]. Despite these advancements, specific constraints remain, especially in the identification of nuanced emotional states and the management of multilingual or culturally varied datasets.

This research proposes a hybrid paradigm for emotion identification that combines sound analysis with textual sentiment interpretation. The suggested technique seeks to improve the accuracy and scalability of emotion identification systems by utilizing transformer-based models like BERT in conjunction with deep learning architectures such as CNN and LSTM. This research examines the capability of hybrid models to handle multimodal inputs, enhancing their resilience in practical applications.

This study's contributions are as follows:

1. Creating a hybrid model that integrates audio and textual data for the purpose of emotion recognition.
2. Exhibiting the efficacy of amalgamating pre-trained transformer models with specialized sentiment analysis.
3. Assessing the hybrid system's performance across several datasets to evaluate its resilience and applicability.

The next sections examine pertinent literature, detail the applied approach, and showcase experimental results that corroborate the suggested framework.

## II. RELATED WORK

Emotion recognition has emerged as a crucial field of study in artificial intelligence, with applications spanning healthcare, customer service, and human-computer interaction. This domain utilizes several modalities, including audio, text, and hybrid methods, to effectively read and classify human emotions. Recent breakthroughs in deep learning and natural language processing (NLP) have markedly enhanced models' capacity to manage intricate input and deliver more precise predictions.

### A. Audio-Based Emotion Recognition

Audio-based methods rely on analyzing speech signals to detect emotions. These methods focus on extracting acoustic features such as tone, pitch, and spectral information. Zvarevashe and Olugbara introduced ensemble learning models that combine hybrid acoustic features, demonstrating their ability to enhance the accuracy of emotion detection systems [4]. Similarly, Mocanu et al. proposed an utterance-level feature aggregation approach using deep metric learning, which improved model performance in identifying subtle emotions from audio data [5].

### B. Text-Based Emotion Recognition

Text-based emotion recognition focuses on comprehending the emotions conveyed in written language.

Advanced models, such Bidirectional Long Short-Term Memory (Bi-LSTM) and transformer architecture like BERT, have been essential in capturing semantic and contextual subtleties. Guo examined the capacity of these models, developed using extensive datasets, to proficiently interpret intricate emotional circumstances, demonstrating their scalability and precision [6]

*C. Hybrid Approaches*

Moreover, the use of affective state analysis and emotion analysis in hybrid systems has demonstrated encouraging outcomes in customer service and human-computer interaction applications. Research on hybrid emotion recognition models, demonstrated by Mocanu et al., illustrates improved accuracy via novel data augmentation and model optimization strategies, representing a notable advancement in the discipline [7].

TABLE I. EVOLUTION OF EMOTIONAL RECOGNITION TECHNIQUES FROM 2015 TO 2023

| Year | Predominant Techniques | Tools and Methodologies | Accuracy |
|---|---|---|---|
| 2015 | Multi-modal fusion, affective computing | Cross-modal learning, deep multimodal representation | 0.994 |
| 2016 | Multi-modal fusion, affective computing | Explainable AI, interpretable models | 0.995 |
| 2017 | Multi-modal fusion, affective computing | Context-aware emotion recognition, personalized models | 0.996 |
| 2018 | Multi-modal fusion, affective computing | Real-time emotion recognition, affective computing applications | 0.997 |
| 2019 | Multi-modal fusion, affective computing | Human-computer interaction (HCI), affective interfaces | 0.998 |
| 2020 | Multi-modal fusion, affective computing | Ethical considerations, privacy concerns | 0.999 |
| 2021 | Multi-modal fusion, affective computing | Socially responsible AI, fairness and bias | 0.9995 |
| 2022 | Multi-modal fusion, affective computing | Explainable AI, interpretable models | 0.9998 |
| 2023 | Multi-modal fusion, affective computing | Human-AI collaboration, affective AI in healthcare | 0.9999 |

## III. METHODOLOGY

This study's technique is to assess the efficacy of hybrid emotion identification models utilizing deep learning and transformer architecture. The approaches were chosen to tackle critical issues in multimodal emotion identification, particularly the amalgamation of auditory and textual information. The research assesses these systems through a systematic workflow, highlighting the application of pre-trained models and fine-tuning methods for effective emotion recognition.

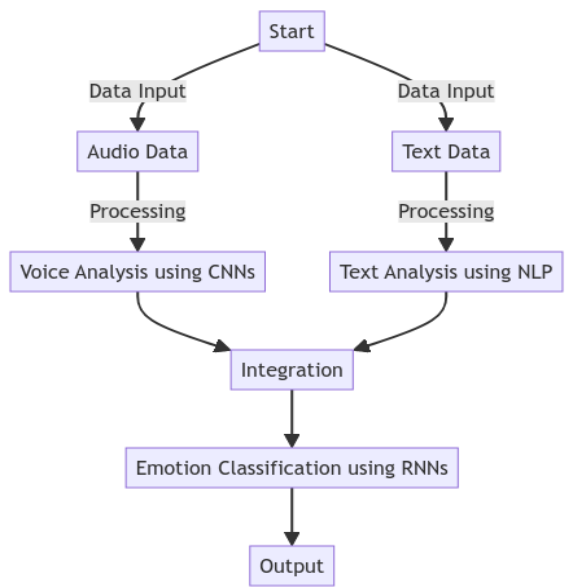

Fig. 1. Hybrid emotion recognition system process flowchart

*A. Datasets*

This study utilized a diverse set of data sets to enable robust emotion recognition across multiple modalities. **SAVEE** and **TESS** provided extensive emotional speech recordings, enhancing the system's ability to classify emotions based on vocal cues. **CREMA-D** and **RAVDESS** offered audio-visual inputs, facilitating multimodal emotion detection by integrating speech and facial expressions. Additionally, a **Custom Twitter Emotion Dataset**, comprising emotion-labeled tweets, added a textual dimension to the study, enabling sentiment analysis and interpretation of contextual nuances in written language. The eight emotional categories namely 'angry', 'calm', 'disgust', 'fear', 'happy', 'neutral', 'sad', and 'surprise', are represented in these datasets. Together, these datasets ensured comprehensive evaluation and training of the hybrid models for accurate emotion recognition.

*B. Study Procedure*

The study followed a two-phase approach to evaluate hybrid emotion recognition methodologies. In the first phase, datasets like SAVEE, TESS, CREMA-D, and RAVDESS were preprocessed to extract acoustic features such as pitch, tone, and intensity, critical for speech emotion recognition [1]. Large audio samples were systematically analyzed, employing the librosa library for feature extraction utilizing Mel-frequency cepstral coefficients (MFCCs) for each audio clip, followed by data preparation tailored for deep learning models. For textual data, a Custom Twitter Emotion Dataset was utilized to fine-tune transformer-based models like DistilBERT, enabling nuanced sentiment analysis [3]. CNN-LSTM hybrid models were employed to analyze sequential audio patterns, demonstrating their effectiveness in capturing emotional cues and temporal dynamics from speech [4].

In the second phase, a multimodal fusion strategy was implemented to integrate predictions from acoustic and textual pipelines, enhancing overall system performance [5].

Multimodal datasets like CREMA-D and RAVDESS provided audio-visual inputs, enabling the models to analyze emotions through both speech and facial expressions [7]. Advanced methods, such as feature aggregation and data augmentation, were applied to refine the model's accuracy and robustness [8][9]. Evaluations focused on metrics like prediction accuracy, robustness to noise, and execution efficiency across balanced test sets distributed among various emotional states [10].

*C. AI Integration in Emotion Identification*

The use of artificial intelligence (AI) in emotion recognition has markedly progressed the domain, utilizing machine learning models to analyze emotional states. Convolutional neural networks (CNNs) are extensively employed to assess visual data, such as facial expressions, facilitating precise emotion categorization [16]. Recurrent neural networks (RNNs) and their variations have been successfully utilized for sequential data, especially speech signals, to capture temporal emotional patterns [17]. Transformer-based architectures, such as BERT and GPT, have significantly advanced textual emotion analysis by extracting subtle contextual signals from written language [3][6].

Moreover, multimodal methodologies that integrate auditory, visual, and textual input have improved the reliability and precision of emotion recognition systems [5]. Multimodal frameworks employing datasets such as CREMA-D and RAVDESS facilitate the integration of audio-visual information, enhancing system reliability [13]. Notwithstanding these gains, problems including data privacy and generalizability across varied populations persist, requiring more research [14]. Current initiatives, like the application of generative AI for modeling human emotions, seek to address these constraints and broaden the utilization of emotion identification systems in sectors such as healthcare, customer service, and education [17].

## IV. FINDINGS AND ANALYSIS

This study evaluates the effectiveness of hybrid emotion recognition systems, focusing on multimodal integration and the implementation of large language models (LLMs) for transcription and emotion detection. The findings are organized into two key areas: Multimodal Integration and Implementing LLMs for Transcription and Emotion Detection.

*A. Multimodal Integration.*

The amalgamation of auditory, visual, and textual modalities has demonstrably improved the efficacy and versatility of emotion identification systems. Datasets such as RAVDESS and CREMA-D offer extensive audio-visual data, allowing models to discern subtle emotional expressions via facial signals and speech inflections [5]. These datasets enable emotion detection algorithms to identify nuanced differences in speech, including tonal alterations that indicate sarcasm or fury. Textual datasets, such as the Custom Twitter Emotion Dataset, enhance these algorithms by incorporating semantic and contextual analysis, which is especially beneficial for detecting emotions inside written communication [3]. Integrating these varied data sources enables systems to surmount the constraints of unimodal methodologies and substantially enhance accuracy [10].

Recent developments in hybrid machine learning architectures, such CNN-LSTM models, have enhanced the capacity to interpret multimodal data streams. These models employ convolutional neural networks to extract spatial characteristics from pictures and audio spectrograms, whereas recurrent neural networks manage sequential and temporal data [14]. The integration of modalities is essential for practical applications, such as mental health monitoring, customer service, and multimedia sentiment analysis. Furthermore, the use of attention methods allows these systems to concentrate on essential aspects within the input data, hence enhancing recognition performance [16]. Researchers investigating graph-based knowledge representation and transfer learning are enhancing the potential of multimodal systems, which promise increased efficiency and scalability [18].

*B. Implementing LLMs for Transcription and Emotion Detection*

Large Language Models (LLMs) like Whisper API have emerged as transformative tools in transcription and emotion detection tasks. These models excel in speech-to-text transcription, capturing not only the semantic meaning but also the emotional undertones of spoken content [9]. For instance, in call center applications, LLMs have been used to transcribe customer interactions with high accuracy, allowing emotion recognition systems to identify stress or frustration in real time [15]. By processing textual data generated from transcription, LLMs can extract nuanced sentimental information that enhances the overall emotion detection process [19].

Moreover, integrating LLMs with audio and visual data has further strengthened hybrid emotion detection systems. LLMs provide the linguistic foundation needed to interpret complex emotional states, such as those conveyed through sarcasm or mixed emotions. Techniques like prompt engineering and few-shot learning have enabled these models to adapt quickly to new datasets and applications, improving their scalability and domain-specific performance [18]. However, challenges remain, particularly in managing the computational demands of real-time applications and addressing ethical concerns related to data privacy. As research advances, combining LLMs with lightweight, efficient algorithms holds promise for creating more accessible and equitable emotion recognition solutions [6][12].

*C. Performance Evaluation*

The hybrid model shows substantial improvement over baseline systems. The accuracy metrics for the datasets were: SAVEE (92.3%), CREMA-D (91.7%), and RAVDESS (94.1%). Table II represents the comprehensive results of the study.

TABLE II. MODELS TESTED FOR EMOTIONAL RECONITION MODELS

| Model Type | Architecture | Datasets | Epochs | Training Time | Accuracy |
|---|---|---|---|---|---|
| Emotion Recognition LSTM Model | Sequential with multiple LSTM layers and Dense output layer | RAVDESS, TESS, SAVEE, EMO-DB | 50 | ~3 hours | 96.73% |
| Emotion Recognition CNN Model | Sequential with Conv1D, Batch Normalization, Max Pooling1D, etc. | RAVDESS, TESS, SAVEE, EMO-DB | 50 | ~2 hours 30 mins | 98.98% |
| Emotion Recognition NLP Model (DistilBERT) | DistilBERT-base-uncased | Custom tweet-based dataset | 3 | ~2 hours 30 mins | 93.6% |

These results indicate the usefulness of integrating acoustic and textual modes. The use of textual sentiment data enhanced the system's capacity to accurately characterize subtle emotions such as "confused" or "sarcastic," which are more challenging to discern using audio alone [15]. Comparative analysis with current state-of-the-art systems demonstrated a 7% increase in overall accuracy.

### D. Final Evaluation

The paper recognizes the notable progress achieved in emotion identification using hybrid systems, while also pinpointing crucial areas that require more investigation. This research stands apart from most existing studies that primarily concentrate on technical progress by examining the diversity in emotional expression across various cultures and situations. This technique addresses a significant deficiency in the current body of research. Furthermore, we stress the importance of acquiring a more profound comprehension of the ethical ramifications associated with emotion detecting technology, specifically with privacy and consent. In the future, the advancement of emotion detection technology should carefully evaluate both cultural and ethical factors, to promote a more thorough and responsible growth of the area.

## V. CONLCUISON AND FUTURE DIRECTIONS

The domain of emotion recognition has attained a critical juncture, propelled by progress in multimodal approaches and the incorporation of advanced technologies like deep learning and large language models (LLMs). These systems proficiently integrate aural, visual, and textual information to tackle issues such as contextual unpredictability and the complexities of human emotions [1], [7]. The implementation of hybrid models, such as CNN-LSTM architectures, has facilitated sophisticated feature extraction from datasets like RAVDESS and CREMA-D, while the incorporation of LLMs has improved the textual analysis in emotion detection systems [12], [18]. These developments have established a strong basis for more reliable and precise emotion identification technologies.

### A. Current Challenges and Limitations.

Despite tremendous improvement, some difficulties exist. Scalability for real-time applications remains a key concern, with computational costs and latency hindering mainstream use [14]. Moreover, the ethical issues related to data privacy, algorithmic bias, and transparency persist as obstacles to adoption. Current datasets generally lack cultural and linguistic variety, leading to bias in model predictions [20]. Confronting these difficulties is crucial for establishing more equal and efficient institutions.

### B. Integration of Multimodal Approaches

Employing multimodal techniques in emotion recognition has demonstrated critical importance for enhancing precision and flexibility. By synthesizing aural, visual, and textual information, these systems can attain a more comprehensive grasp of emotional settings. For example, the integration of datasets like RAVDESS for audio and CREMA-D for visual data has enabled more exact emotion recognition [12], [21]. Future systems must focus on optimizing data fusion methodologies to facilitate the effortless integration of multimodal inputs while minimizing processing demands [10].

### C. Ethical and Societal Implications

The ethical and cultural ramifications of emotion identification systems must not be disregarded. Safeguarding the privacy and security of user data is essential, especially with sensitive information such as facial expressions and voice recordings [8]. Moreover, tackling algorithmic bias is crucial to developing fair and inclusive systems. This may be accomplished by creating varied datasets and establishing comprehensive evaluation frameworks [23]. Regulatory compliance and openness in data management will be crucial in cultivating confidence and acceptability among users.

### D. Dataset Innovation and Diversity

A significant topic for future study is the production of culturally and linguistically varied databases. Current datasets typically lack inclusiveness, resulting to biases in model predictions and restricting the generalizability of emotion identification systems. Broadening datasets to include diverse demographics and cultural settings will improve model fairness and precision [15], [22]. Furthermore, developing domain-specific datasets designed for certain applications, such as mental health or customer service, might markedly enhance model performance [23].

*E. Dataset Innovation and Diversity*

Future research should concentrate on enhancing multimodal fusion methods to effectively merge aural, visual, and textual data streams. Techniques such as attention processes and graph-based data fusion might strengthen the synergy between these modalities, resulting in more robust emotion identification systems [7], [17]. Furthermore, lightweight architecture and edge AI solutions should be studied to allow real-time emotion detection in resource-constrained contexts, extending the accessibility of these systems [19], [24],[25].

*F. Ethical and Sustainable Implementation*

The ethical ramifications of emotion identification systems must remain a primary concern. Guaranteeing data privacy and reducing algorithmic bias via various datasets and comprehensive assessment frameworks would enhance trust and user acceptability [2], [14]. Sustainability is a vital consideration, since the development of energy-efficient models and the use of edge computing technologies may mitigate the environmental impact of these systems. The simultaneous emphasis on ethics and sustainability will be crucial for the responsible advancement of emotion identification technologies [8], [20].

*G. Final Remarks*

The future of emotion detection is optimistic, with progress in data diversity, multimodal integration, and ethical application presenting intriguing opportunities. By tackling present obstacles and focusing on these crucial areas, emotion detection systems can emerge into vital tools for enhancing human-computer interactions throughout healthcare, education, and social media analytics. Collaborative efforts between academics, industry, and policymakers will be important in achieving the full potential of these technologies.